\documentclass[10pt,twocolumn,letterpaper]{article}

\usepackage[applications]{wacv} 
 \usepackage{multirow}
 \usepackage[accsupp]{axessibility}

\definecolor{wacvblue}{rgb}{0.21,0.49,0.74}
\usepackage[pagebackref,breaklinks,colorlinks,allcolors=wacvblue]{hyperref}

\def\wacvPaperID{*****} 
\def\confName{WACV}
\def\confYear{2026}

\title{FUME: Fused Unified Multi-Gas Emission Network for Livestock Rumen Acidosis Detection}

\author{Taminul Islam\textsuperscript{1}, Toqi Tahamid Sarker\textsuperscript{1}, Mohamed Embaby\textsuperscript{2}, Khaled R Ahmed\textsuperscript{1}, Amer AbuGhazaleh\textsuperscript{1}\\
\textsuperscript{1}Southern Illinois University, Carbondale, \textsuperscript{2}University of California, Davis\\
{\tt\small \{taminul.islam, toqitahamid.sarker, khaled.ahmed, aabugha\}@siu.edu}, \tt\small membaby@ucdavis.edu
}

\begin{document}
\maketitle
\begin{abstract}
Ruminal acidosis is a prevalent metabolic disorder in dairy cattle causing significant economic losses and animal welfare concerns. Current diagnostic methods rely on invasive pH measurement, limiting scalability for continuous monitoring. We present FUME (Fused Unified Multi-gas Emission Network), the first deep learning approach for rumen acidosis detection from dual-gas optical imaging under in vitro conditions. Our method leverages complementary carbon dioxide (CO$_2$) and methane (CH$_4$) emission patterns captured by infrared cameras to classify rumen health into Healthy, Transitional, and Acidotic states. FUME employs a lightweight dual-stream architecture with weight-shared encoders, modality-specific self-attention, and channel attention fusion, jointly optimizing gas plume segmentation and classification of dairy cattle health. We introduce the first dual-gas OGI dataset comprising 8,967 annotated frames across six pH levels with pixel-level segmentation masks. Experiments demonstrate that FUME achieves 80.99\% mIoU and 98.82\% classification accuracy while using only 1.28M parameters and 1.97G MACs—outperforming state-of-the-art methods in segmentation quality with 10$\times$ lower computational cost. Ablation studies reveal that CO$_2$ provides the primary discriminative signal and dual-task learning is essential for optimal performance. Our work establishes the feasibility of gas emission-based livestock health monitoring, paving the way for practical, in vitro acidosis detection systems. Codes are available at \url{https://github.com/taminulislam/fume}.
\end{abstract}

\section{Introduction}
\label{sec:introduction}

Ruminal acidosis represents one of the most economically significant metabolic disorders in modern dairy and beef cattle production, affecting an estimated 19--26\% of dairy cows in intensive farming systems worldwide \cite{plaizier2008subacute,voulgarakis2024subacute}. This condition, characterized by abnormally low rumen pH (typically below 5.8), results from the accumulation of volatile fatty acids, lactic acid in particular, during rapid fermentation of high-concentrate diets. The consequences extend beyond immediate digestive dysfunction: chronic acidosis leads to reduced feed intake, decreased milk production (losses of 1--3 kg/day), laminitis, liver abscesses, systemic inflammation, and in severe cases, mortality \cite{owens1998acidosis}. Subacute ruminal acidosis costs the North American dairy industry an estimated \$500 million to \$1 billion annually, with per-cow losses of approximately \$400--475 per lactation \cite{enemark2008monitoring}, figures that likely underestimate current losses given inflation and increased production intensity \cite{voulgarakis2024can}.

Early detection of acidosis is critical for timely intervention and prevention of irreversible health damage. However, current diagnostic methods present significant practical limitations. The gold standard---direct rumen pH measurement via rumenocentesis or indwelling bolus sensors---is invasive, expensive, requires veterinary intervention, and causes animal stress \cite{gantner2025prevalence}. Indirect indicators such as milk composition, fecal pH, and behavioral monitoring lack the sensitivity and specificity required for reliable early-stage detection \cite{christodoulopoulos2025subacute}. Consequently, acidosis often remains undiagnosed until clinical signs become apparent, by which point substantial metabolic damage has already occurred. The agricultural industry urgently needs real-time and cost-effective tools for continuous rumen health monitoring at scale.

Rumen fermentation produces two primary gaseous products: carbon dioxide (CO\textsubscript{2}) and methane (CH\textsubscript{4}), whose production rates and relative proportions are directly influenced by rumen pH and microbial activity \cite{janssen2010influence}. Under acidotic conditions, methanogenic archaea activity is suppressed, reducing CH\textsubscript{4} production, while CO\textsubscript{2} generation from acidic fermentation pathways increases \cite{mao2025rumen}. These characteristic gas emission patterns, captured by infrared optical gas imaging (OGI) cameras, encode valuable information about underlying rumen health status. However, translating raw thermal imagery into actionable health diagnostics requires sophisticated pattern recognition beyond human visual interpretation---a challenge ideally suited to deep learning.

Before deploying gas emission monitoring systems in complex farm environments with variable conditions, free-moving animals, and multiple confounding factors, it is essential to first validate the fundamental feasibility of pH prediction from gas signatures under controlled conditions. In vitro fermentation systems, such as the ANKOM RF Gas Production System \cite{embaby2025optical}, provide precisely regulated pH environments with continuous gas monitoring, enabling the establishment of ground-truth relationships between pH levels and observable gas emission patterns. This controlled validation is critical for several reasons: in vitro systems eliminate confounding variables present in live animals---diet variation, individual metabolic differences, stress responses, and environmental fluctuations---allowing isolation of the pH-gas emission relationship. They also enable systematic collection of data across the full pH spectrum, including severe acidotic states (pH $<$ 5.0) that would be unethical to induce in live animals. Furthermore, they provide the labeled training data necessary for supervised deep learning, as ground-truth pH measurements can be precisely controlled and verified.

\begin{figure}[t]
\centering
\includegraphics[width=\columnwidth]{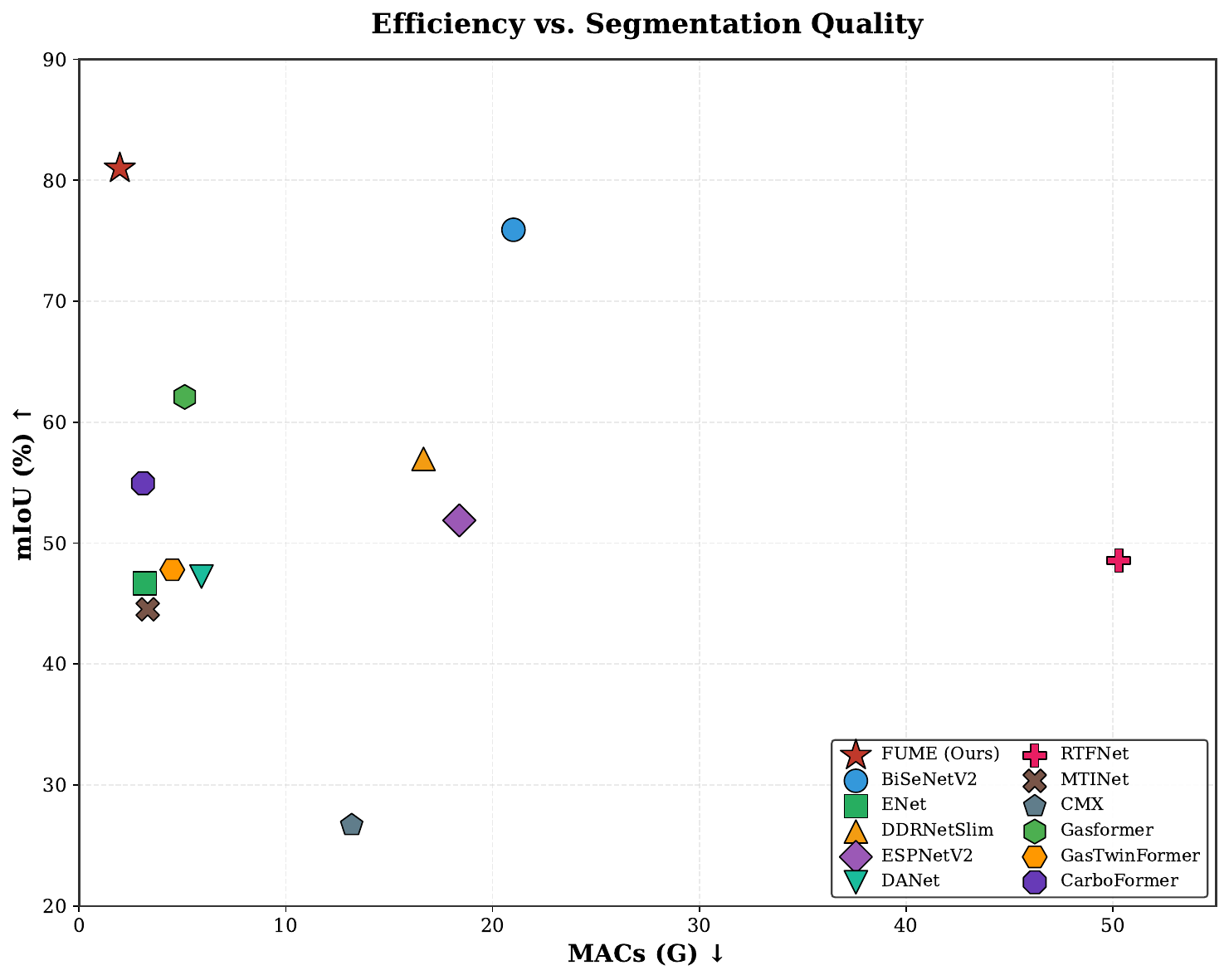}
\caption{Efficiency vs. segmentation quality trade-off across all compared methods including OGI-specific baselines. The upper-left region represents optimal performance (high mIoU, low computational cost). FUME achieves the best trade-off, delivering state-of-the-art segmentation quality while requiring significantly fewer MACs than competing methods.}
\label{fig:efficiency}
\end{figure}

In this paper, we present the first deep learning approach for rumen acidosis detection from dual-gas optical imaging under controlled laboratory conditions. Our method leverages the complementary information encoded in simultaneous CO\textsubscript{2} and CH\textsubscript{4} emissions to predict rumen health status across three clinically relevant categories: Healthy, Transitional, and Acidotic. We formulate this as a multi-task learning problem, jointly optimizing pixel-level gas plume segmentation and image-level livestock health classification. The segmentation task forces the model to learn spatially precise representations of gas distributions, which in turn improves classification by focusing on emission-specific features rather than irrelevant background or equipment artifacts. Our dual-stream architecture processes CO\textsubscript{2} and CH\textsubscript{4} channels independently before fusing them via channel attention, enabling the model to learn gas-specific patterns as well as their interactions. As illustrated in Fig.~\ref{fig:efficiency}, FUME achieves the best accuracy-efficiency trade-off among all compared methods, occupying the optimal upper-left region with highest segmentation quality at minimal computational cost.

The main contributions of this work are twofold: (1) we introduce the first dual-gas OGI dataset for rumen health assessment, comprising 8,967 annotated frame pairs (CO\textsubscript{2} and CH\textsubscript{4}) across six pH levels (Table~\ref{tab:dataset_stats}), with pixel-level segmentation masks and image-level health labels---a novel resource for the agricultural computer vision community; and (2) we propose a dual-stream multi-task deep learning architecture that jointly performs gas plume segmentation and health classification, establishing the feasibility of rumen pH inference from gas emission patterns under in vitro conditions and paving the way for practical, scalable livestock health monitoring systems.

\section{Related Work}
\label{sec:related}

Our work intersects three research areas: gas detection and environmental monitoring, multi-modal fusion and multi-task learning for vision tasks, and agricultural computer vision. We review recent advances in each domain and position our contributions relative to the state-of-the-art.

\begin{figure*}[t]
\centering
\includegraphics[width=0.85\textwidth]{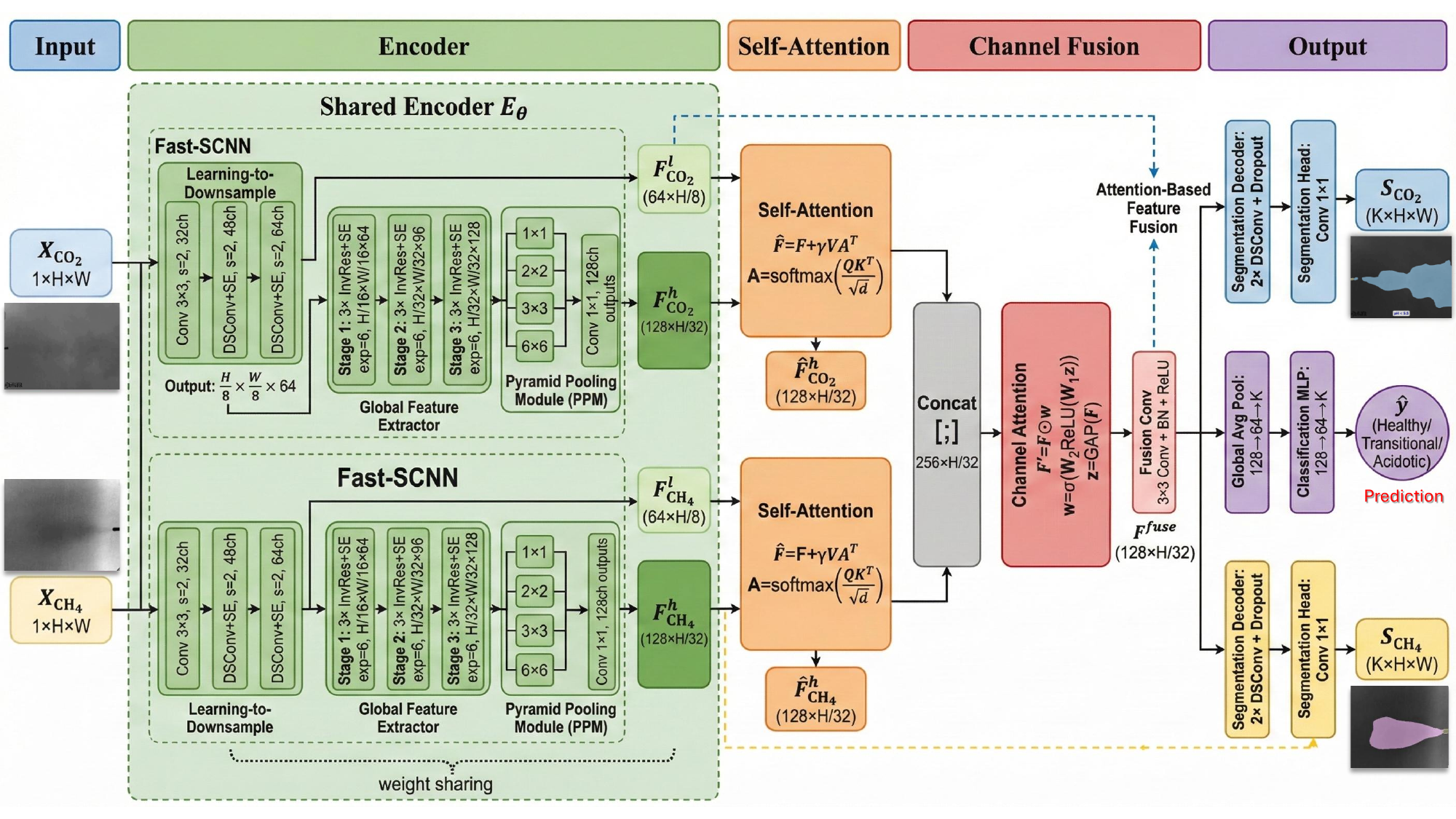}
\caption{\textbf{FUME Architecture.} Our dual-stream network processes paired CO$_2$ and CH$_4$ thermal images through a weight-shared Fast-SCNN encoder with depthwise separable convolutions. The Learning-to-Downsample module reduces spatial resolution to $\frac{H}{8} \times \frac{W}{8}$, while the Global Feature Extractor with Pyramid Pooling Module captures multi-scale context at $\frac{H}{32} \times \frac{W}{32}$. Modality-specific self-attention modules refine each stream before channel attention fusion combines the representations. The fused features feed dual segmentation decoders and a classification head.}
\label{fig:fume-model}
\end{figure*}

\noindent\textbf{Gas Detection and Environmental Monitoring.}
Recent work has addressed gas plume segmentation in OGI domains. Gasformer~\cite{sarker2024gasformer} proposes lightweight transformer hybrids for CH$_4$ detection with attention to boundary quality. GasTwinFormer~\cite{tahamid2025gastwinformer} combines segmentation with classification in multi-task OGI. CarboFormer~\cite{islam2025carbonext} addresses CO$_2$ plume segmentation. Other progress includes methane leak detection~\cite{lu2025vehicle}, wildfire smoke detection~\cite{wang2024rfwnet}, and atmospheric pollutant estimation~\cite{sohrabi2025fusing}. While these methods target single-gas industrial scenarios, FUME addresses dual-gas fusion for metabolic state inference---a fundamentally different problem requiring cross-modal reasoning for health assessment rather than binary presence detection.

\noindent\textbf{Multi-Modal Fusion and Multi-Task Learning.}
Multi-modal fusion has proven effective for tasks requiring complementary information sources across diverse vision applications. Cross-modal attention mechanisms for RGB-thermal fusion in semantic segmentation demonstrate that learned attention outperforms fixed fusion rules \cite{fu2025cafnet}---a principle we adopt for CO\textsubscript{2}-CH\textsubscript{4} fusion. Hierarchical fusion approaches for RGB-D salient object detection show that multi-scale aggregation improves robustness to modality-specific artifacts \cite{chen2018progressively}, while dual-stream architectures for multi-modal tumor segmentation validate the effectiveness of parallel encoders with late fusion for heterogeneous inputs \cite{hu2025dual}. Adaptive fusion networks that dynamically weight modality contributions based on input quality \cite{cheng2024adaptive} and comparisons of feature-level versus decision-level fusion \cite{wei2025hierarchical} provide insights that inform our fusion module design. Concurrently, multi-task learning (MTL) has emerged as a powerful paradigm for leveraging task synergies, with end-to-end MTL frameworks for autonomous driving \cite{xing2025openemma}, task-balanced learning strategies addressing conflicting gradients \cite{liu2025mmtl}, joint segmentation and classification of medical lesions demonstrating complementary supervision signals \cite{li2024ssm}, multi-task attention networks for scene understanding \cite{xu2024unified}, and efficient MTL architectures for resource-constrained deployment \cite{sampath2025mtl}. Despite these advances, existing fusion and MTL approaches target RGB, depth, or thermal modalities rather than multi-gas emissions with distinct physical meanings, leaving the domain of gas-based health inference unexplored.

\noindent\textbf{Agricultural Computer Vision.}
Computer vision for livestock monitoring has focused primarily on external observations, including deep learning for behavior recognition \cite{zhang2024time}, automated body condition scoring \cite{angel2024comparison}, lameness detection from gait analysis \cite{li2024novel}, and vision-based health monitoring systems \cite{jiao2024vision}. In crop and weed monitoring, researchers have developed precision agriculture systems using multi-spectral imagery for disease detection \cite{ngugi2024revolutionizing} and weed segmentation with hierarchical vision transformers \cite{islam2025weedswin,sarker2025weedsense}. While attention mechanisms have revolutionized computer vision through cross-modal learning \cite{kim2023cross} and improved transformer architectures for dense prediction \cite{tang2023swinlstm}, their data requirements make them less suitable for limited annotated data---motivating our CNN-based approach with targeted attention modules. Critically, all existing agricultural vision methods observe external animal characteristics and cannot directly assess internal metabolic states such as rumen pH.

\begin{table*}[t]
\centering
\caption{Quantitative comparison with state-of-the-art methods. $\dagger$ indicates best performance. For HD95, ASD, and Latency, lower is better.}
\label{tab:main_results}
\resizebox{\textwidth}{!}{
\begin{tabular}{l|c|cc|cc|cccc}
\toprule
\multirow{2}{*}{\textbf{Model}} & \textbf{Classification} & \multicolumn{2}{c|}{\textbf{Segmentation}} & \multicolumn{2}{c|}{\textbf{Boundary}} & \multicolumn{4}{c}{\textbf{Efficiency}} \\
\cmidrule(lr){2-2} \cmidrule(lr){3-4} \cmidrule(lr){5-6} \cmidrule(lr){7-10}
& Acc (\%) & mIoU (\%) & Dice (\%) & HD95 (px) & ASD (px) & Params (M) & MACs (G) & Latency (ms) & FPS \\
\midrule
BiSeNetV2~\cite{yu2021bisenet} & \textbf{99.04}$^\dagger$ & 75.91 & 85.85 & 72.13 & 20.89 & 2.77 & 21.01 & 3.54 & 282.49 \\
ENet~\cite{paszke2016enet} & 97.86 & 46.67 & 54.13 & 101.16 & 34.24 & \textbf{0.26}$^\dagger$ & 3.18 & 2.71 & 369.00 \\
DDRNetSlim~\cite{hong2021deep} & 94.98 & 56.97 & 70.55 & 140.50 & 46.13 & 3.39 & 16.66 & \textbf{1.66}$^\dagger$ & \textbf{602.41}$^\dagger$ \\
ESPNetV2~\cite{mehta2019espnetv2} & 90.81 & 51.88 & 58.00 & 92.32 & 26.55 & 2.22 & 18.39 & 2.58 & 387.60 \\
DANet~\cite{fu2019dual} & 48.50 & 47.23 & 54.49 & 103.85 & 34.65 & 1.43 & 5.92 & 2.99 & 334.45 \\
RTFNet~\cite{putra2025residual} & 46.47 & 48.56 & 55.55 & 107.90 & 33.22 & 35.01 & 50.28 & 3.88 & 257.73 \\
MTINet~\cite{vandenhende2020mti} & 46.37 & 44.53 & 52.27 & 113.11 & 39.20 & 1.80 & 3.32 & 3.18 & 314.47 \\
CMX~\cite{zhang2023cmx} & 33.65 & 26.69 & 29.64 & 187.34 & 61.28 & 4.41 & 13.18 & 1.98 & 505.05 \\
Gasformer~\cite{sarker2024gasformer} & 89.56 & 62.09 & 74.31 & 42.51 & 10.56 & 3.65 & 5.11 & 5.63 & 177.50 \\
GasTwinFormer~\cite{tahamid2025gastwinformer} & 86.27 & 47.80 & 55.77 & 47.77& 14.18 & 3.35 & 4.51 & 6.32 & 158.20 \\
CarboFormer~\cite{islam2025carbonext} & 87.92 & 54.95 & 65.04 & 41.14 & 10.37 & 3.81 & 3.09 & 9.31 & 107.40 \\
\midrule
\textbf{FUME (Ours)} & 98.82 & \textbf{80.99}$^\dagger$ & \textbf{89.23}$^\dagger$ & \textbf{46.58}$^\dagger$ & \textbf{14.00}$^\dagger$ & 1.28 & \textbf{1.97}$^\dagger$ & 3.06 & 326.80 \\
\bottomrule
\end{tabular}
}
\end{table*}

\noindent\textbf{Research Gap.}
Despite significant advances in gas detection, multi-modal fusion, and agricultural computer vision, a critical gap remains: no existing work has explored the use of gas emission imaging for livestock health assessment. Industrial gas detection systems focus on binary presence detection rather than physiological state inference. Agricultural vision systems monitor external observations but cannot assess internal metabolic conditions. Our work bridges this gap by combining dual-gas fusion with multi-task learning specifically for rumen health monitoring under in vitro conditions, demonstrating that CO\textsubscript{2} and CH\textsubscript{4} emission patterns captured through OGI technology encode sufficient information to distinguish healthy, transitional, and acidotic states---establishing proof-of-concept for gas-based metabolic assessment. Specifically, we address the efficiency limitations of existing multi-modal fusion approaches by proposing channel attention fusion instead of computationally expensive cross-modal attention, achieving state-of-the-art segmentation quality with significantly reduced computational cost suitable for real-time edge deployment.

\section{Method}
\label{sec:method}

We present FUME (Fused Unified Multi-gas Emission Network), a lightweight dual-stream architecture for acidosis detection from paired CO$_2$ and CH$_4$ thermal emissions. While individual components (Fast-SCNN, self-attention, channel attention) are established, our key contributions are: (1) \textit{novel problem formulation}---first to demonstrate that dual-gas emission patterns encode metabolic state, enabling non-invasive health inference; (2) \textit{physiologically-motivated fusion}---channel attention over cross-attention based on our finding that CO$_2$ and CH$_4$ lack pixel-wise correspondence but exhibit correlated aggregate statistics; and (3) \textit{multi-task synergy}---joint segmentation and classification where spatial precision improves classification focus.

\subsection{Problem Formulation}

Given paired thermal imaging frames $\mathbf{X}_{\text{CO}_2}, \mathbf{X}_{\text{CH}_4} \in \mathbb{R}^{1 \times H \times W}$ capturing CO$_2$ and CH$_4$ gas emissions, our objective is to jointly predict: (i) pixel-wise segmentation masks $\mathbf{M}_{\text{CO}_2}, \mathbf{M}_{\text{CH}_4} \in \{0,1,2\}^{H \times W}$ identifying background, tube, and gas emission regions, and (ii) a global classification label $y \in \{\text{Healthy}, \text{Transitional}, \text{Acidotic}\}$ indicating the metabolic acidosis state. The fundamental challenge lies in the \textbf{cross-modal semantic gap}: while CO$_2$ and CH$_4$ concentrations are physiologically correlated, their spatial distributions differ substantially. We propose a solution that captures inter-modal dependencies through efficient channel-wise attention while preserving intra-modal spatial structure via self-attention.

\subsection{Architecture Overview}

FUME comprises four components: (1) a weight-shared Fast-SCNN \cite{poudel2019fast} encoder, (2) modality-specific self-attention modules, (3) a channel attention fusion mechanism, and (4) dual-task decoders. Fig.~\ref{fig:fume-model} illustrates the architecture. The forward pass is:
\begin{equation}
\mathbf{F}^l_m, \mathbf{F}^h_m = \mathcal{E}_\theta(\mathbf{X}_m), \quad \hat{\mathbf{F}}^h_m = \text{SelfAttn}_m(\mathbf{F}^h_m)
\end{equation}
\begin{equation}
\mathbf{F}^{\text{fuse}} = \text{ChannelAttn}([\hat{\mathbf{F}}^h_{\text{CO}_2}; \hat{\mathbf{F}}^h_{\text{CH}_4}])
\end{equation}
where $\mathbf{F}^l_m \in \mathbb{R}^{64 \times \frac{H}{8} \times \frac{W}{8}}$ and $\mathbf{F}^h_m \in \mathbb{R}^{128 \times \frac{H}{32} \times \frac{W}{32}}$ denote low-level and high-level features respectively.

\subsection{Dual-Stream Encoder and Fusion}

\noindent\textbf{Shared Encoder.} A critical design decision is weight sharing between CO$_2$ and CH$_4$ encoder streams. This is motivated by two observations: (i) both thermal modalities share common low-level visual primitives (edges, textures, intensity gradients) that benefit from joint representation learning, and (ii) weight sharing reduces encoder parameters by 50\%, enabling deployment on resource-constrained devices. We adopt Fast-SCNN~\cite{poudel2019fast} as our backbone, which employs a learning-to-downsample module achieving $8\times$ downsampling through depthwise separable convolutions:
\begin{equation}
\mathbf{F}^l = \text{DSConv}_{64}(\text{DSConv}_{48}(\text{Conv}_{32}(\mathbf{X})))
\end{equation}
The computational efficiency of depthwise separable convolutions provides approximately $8\times$ reduction compared to standard convolutions. The global feature extractor applies inverted residual blocks with expansion ratio 6 across three bottleneck stages, progressively extracting semantic features at $\frac{1}{16}$ and $\frac{1}{32}$ resolution. A Pyramid Pooling Module aggregates multi-scale context through adaptive pooling at scales $\{1, 2, 3, 6\}$, enabling the network to capture both local gas emission patterns and global distribution characteristics.

\begin{table}[t]
\centering
\caption{Dataset distribution across pH levels, health classes, and splits.}
\label{tab:dataset_stats}
\setlength{\tabcolsep}{3pt}
\footnotesize
\begin{tabular}{@{}lcccccccc@{}}
\toprule
\multirow{2}{*}{\textbf{pH}} & \multirow{2}{*}{Class} & \multirow{2}{*}{Total} & \multirow{2}{*}{CO$_2$} & \multirow{2}{*}{CH$_4$} & Train & Val & Test \\ 
& & & & & (70\%) & (15\%) & (15\%) \\
\midrule
6.5 & Healthy & 1,008 & 693 & 315 & 706 & 151 & 151 \\
6.2 & Healthy & 1,050 & 721 & 329 & 735 & 158 & 157 \\
5.9 & Transitional & 945 & 651 & 294 & 662 & 142 & 141 \\
5.6 & Acidotic & 3,087 & 2,121 & 966 & 2,161 & 463 & 463 \\
5.3 & Acidotic & 1,512 & 1,039 & 473 & 1,058 & 227 & 227 \\
5.0 & Acidotic & 1,365 & 938 & 427 & 955 & 205 & 205 \\
\midrule
\multicolumn{2}{l}{\textbf{Total}} & 8,967 & 6,163 & 2,804 & 6,277 & 1,346 & 1,344 \\
\bottomrule
\end{tabular}
\end{table}

\noindent\textbf{Self-Attention.} At the deepest feature level ($\frac{H}{32} \times \frac{W}{32}$), we apply self-attention independently to each modality stream. This addresses a fundamental limitation of convolutional encoders: while convolutions excel at capturing local patterns, gas emission analysis requires modeling long-range spatial dependencies---a diffuse gas plume may span the entire field of view, and its shape provides diagnostic information that local receptive fields cannot capture:
\begin{equation}
\mathbf{A}_m = \text{softmax}\left(\frac{\mathbf{Q}_m \mathbf{K}_m^\top}{\sqrt{d_k}}\right), \quad \hat{\mathbf{F}}^h_m = \mathbf{F}^h_m + \gamma_m \cdot \mathbf{V}_m \mathbf{A}_m^\top
\end{equation}
where $\gamma_m$ is a learnable scalar initialized to zero for training stability. Critically, we apply self-attention \textit{separately} to each modality rather than cross-modal attention. This decision is grounded in our empirical finding that CO$_2$ and CH$_4$ emissions, while correlated in aggregate statistics, do not exhibit pixel-wise spatial correspondence---their relationship is better captured through channel-wise statistics.

\noindent\textbf{Channel Attention Fusion.} After self-attention refinement, we concatenate modality-specific features and apply channel attention to model inter-gas dependencies. Unlike spatial cross-attention which asks ``which CO$_2$ locations correspond to which CH$_4$ locations,'' channel attention asks ``which feature channels from CO$_2$ should be combined with which channels from CH$_4$''---a more appropriate inductive bias for our task:
\begin{equation}
\mathbf{z} = \text{GAP}([\hat{\mathbf{F}}^h_{\text{CO}_2}; \hat{\mathbf{F}}^h_{\text{CH}_4}]), \quad \mathbf{w} = \sigma(\mathbf{W}_2 \cdot \text{ReLU}(\mathbf{W}_1 \cdot \mathbf{z}))
\end{equation}
\begin{equation}
\mathbf{F}^{\text{fuse}} = \text{Conv}_{3\times3}([\hat{\mathbf{F}}^h_{\text{CO}_2}; \hat{\mathbf{F}}^h_{\text{CH}_4}] \odot \mathbf{w})
\end{equation}
where $\mathbf{W}_1, \mathbf{W}_2$ form a bottleneck MLP with reduction ratio 16. Global average pooling aggregates statistics over the entire spatial extent, enabling the network to learn correlations like ``high mean CO$_2$ intensity combined with low CH$_4$ variance indicates acidotic state.''

\begin{table}[t]
\centering
\caption{Per-class F1 scores (\%) and balanced accuracy.}
\label{tab:perclass}
\resizebox{\columnwidth}{!}{
\begin{tabular}{l|ccc|cc}
\toprule
\textbf{Model} & \textbf{Healthy} & \textbf{Trans.} & \textbf{Acidotic} & \textbf{Macro F1} & \textbf{Bal. Acc} \\
\midrule
BiSeNetV2~\cite{yu2021bisenet} & \textbf{99.04} & \textbf{94.59} & 99.32 & \textbf{97.65} & \textbf{99.24} \\
ENet~\cite{paszke2016enet} & 97.91 & 80.46 & 99.14 & 92.50 & 98.32 \\
DDRNetSlim~\cite{hong2021deep} & 93.96 & 69.31 & 97.75 & 87.01 & 96.01 \\
ESPNetV2~\cite{mehta2019espnetv2} & 87.80 & 81.97 & 93.18 & 87.65 & 85.71 \\
DANet~\cite{fu2019dual} & 55.90 & 75.61 & 33.99 & 55.17 & 68.41 \\
RTFNet~\cite{putra2025residual} & 55.70 & 0.87 & 33.99 & 29.90 & 40.16 \\
MTINet~\cite{vandenhende2020mti} & 55.65 & 1.12 & 33.76 & 29.80 & 40.10 \\
CMX~\cite{zhang2023cmx} & 50.36 & 1.23 & 1.45 & 16.79 & 33.33 \\
Gasformer~\cite{sarker2024gasformer} & 88.00 & 82.50 & 88.00 & 86.16 & 89.56 \\
GasTwinFormer~\cite{tahamid2025gastwinformer} & 86.00 & 79.00 & 87.00 & 83.98 & 86.27 \\
CarboFormer~\cite{islam2025carbonext} & 87.00 & 81.00 & 87.20 & 85.07 & 87.92 \\
\midrule
\textbf{FUME (Ours)} & 98.72 & 94.59 & \textbf{99.15} & 97.49 & 99.03 \\
\bottomrule
\end{tabular}
}
\end{table}

\subsection{Dual-Task Decoding and Training}

\noindent\textbf{Segmentation.} Separate decoding heads for CO$_2$ and CH$_4$ employ Feature Fusion Modules combining high-level features with low-level encoder features via skip connections:
\begin{equation}
\mathbf{S}_m = \text{Classifier}(\text{FFM}(\mathbf{F}^l_m, \hat{\mathbf{F}}^h_m))
\end{equation}
The classifier consists of depthwise separable convolutions followed by a $1\times1$ projection to $K=3$ classes.

\noindent\textbf{Classification.} Global pH state prediction aggregates $\mathbf{F}^{\text{fuse}}$ via adaptive average pooling through a two-layer MLP with dropout.

\noindent\textbf{Training Objective.} We jointly optimize through multi-task loss:
\begin{equation}
\mathcal{L} = \mathcal{L}_{\text{seg}}^{\text{CO}_2} + \mathcal{L}_{\text{seg}}^{\text{CH}_4} + \lambda \mathcal{L}_{\text{cls}}
\end{equation}
where each $\mathcal{L}_{\text{seg}}^m$ combines Focal loss and Dice loss ($0.5 \cdot \mathcal{L}_{\text{focal}} + 0.5 \cdot \mathcal{L}_{\text{dice}}$) to handle class imbalance and boundary precision, $\mathcal{L}_{\text{cls}}$ is Focal loss with class weights, and $\lambda = 0.5$. Self-attention uses single-head with $d_k = C/8 = 16$ and learnable $\gamma$ initialized to zero; channel attention uses reduction ratio 16. FUME achieves 1.28M parameters and 1.97G MACs as seen in Fig.~\ref{fig:efficiency} through weight sharing and efficient fusion.

\begin{figure*}[t]
\centering
\includegraphics[width=0.85\textwidth]{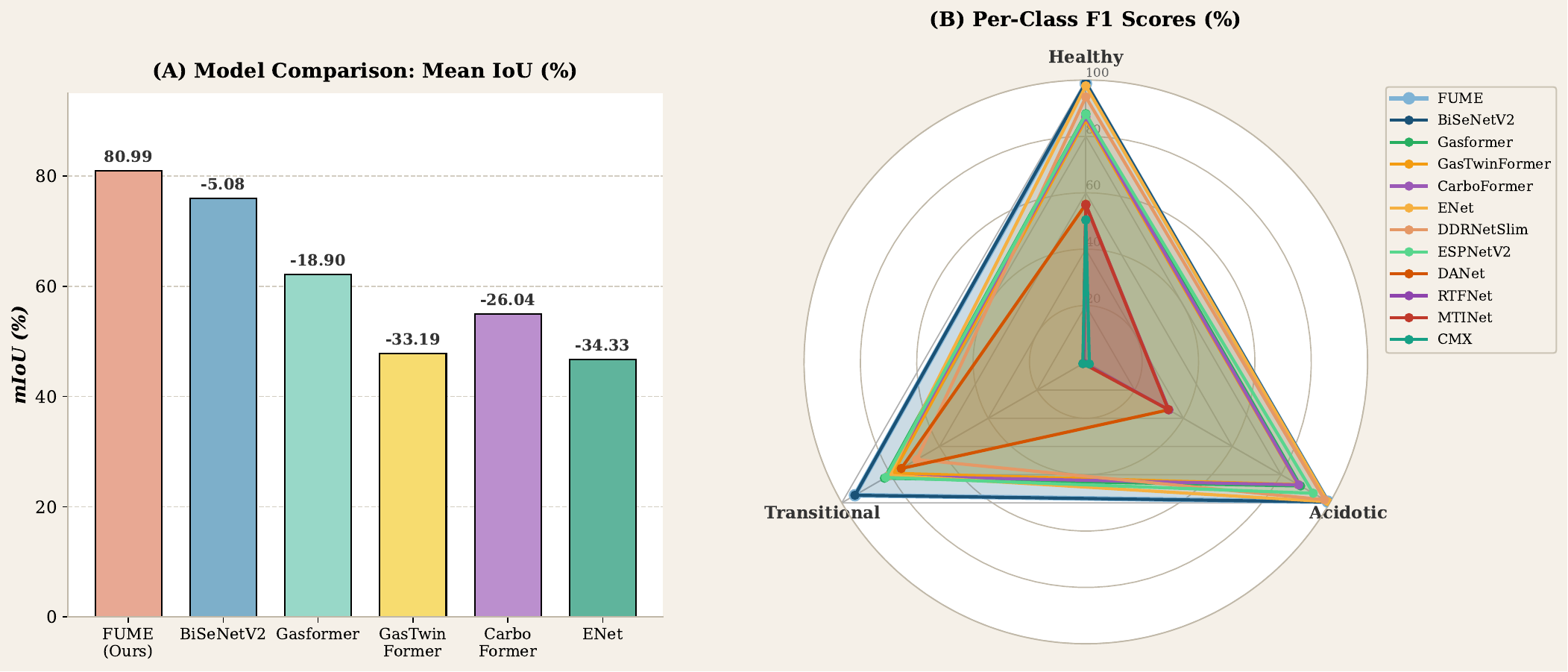}
\caption{Comparison with state-of-the-art methods. (A) Mean IoU comparison showing FUME substantially outperforms both general segmentation baselines and OGI-specific methods (Gasformer, GasTwinFormer, CarboFormer). (B) Per-class F1 radar chart demonstrating balanced classification performance across Healthy, Transitional, and Acidotic classes, where FUME maintains consistent high performance while several baselines exhibit severe degradation on challenging classes.}
\label{fig:comparison_combined}
\end{figure*}

\section{Experiments}
\label{sec:experiments}

\subsection{Dataset}
\label{subsec:dataset}

Rumen fluid was collected from fistulated cattle following protocols, and fermentation substrates were prepared using standard in vitro rumen fermentation procedures with the ANKOM \cite{embaby2019effect}. Samples were maintained at six controlled pH levels (6.5, 6.2, 5.9, 5.6, 5.3, and 5.0), representing the spectrum from severe acidosis to healthy rumen conditions. We captured gas emissions using two specialized FLIR OGI cameras: the FLIR GF343 for CO$_2$ detection (4.2--4.4 $\mu$m mid-wave infrared) \cite{islam2025carbonext}, and the FLIR GF320 for CH$_4$ detection (3.2--3.4 $\mu$m) \cite{tahamid2025gastwinformer}. Both cameras were co-located at 20~cm from the gas source, recording simultaneously at 256$\times$256 resolution; frames are bilinearly upsampled to 512$\times$512 for training.

We map pH levels to three clinically relevant health classes based on established rumen physiology \cite{plaizier2008subacute}: \textbf{Healthy} (pH $\geq$ 6.0), \textbf{Transitional} (5.8 $\leq$ pH $<$ 6.0), and \textbf{Acidotic} (pH $<$ 5.8). Since gas emissions are intermittent, we extracted only frames containing visible gas plumes and applied data augmentation. Each frame was manually annotated with pixel-level masks ($K$=3 classes: background, tube, gas), with gas plumes occupying only 5.3\% of pixels. Table~\ref{tab:dataset_stats} shows the dataset distribution with stratified 70-15-15 splits; data was partitioned by pH level to prevent temporal leakage. Additionally, frames from the same fermentation session were kept within a single split to ensure independence between training and evaluation sets. The unequal CO$_2$ vs CH$_4$ counts reflect physiological differences: CH$_4$ emissions are less visible at acidotic pH. Unpaired samples use zero-padding with a modality mask indicating available inputs.

\subsection{Baseline Models and Implementation Details}
\label{subsec:baselines}

We compare FUME against state-of-the-art efficient segmentation architectures including BiSeNetV2~\cite{yu2021bisenet}, ENet~\cite{paszke2016enet}, DDRNetSlim~\cite{hong2021deep}, ESPNetV2~\cite{mehta2019espnetv2}, DANet~\cite{fu2019dual}, RTFNet~\cite{putra2025residual}, MTINet~\cite{vandenhende2020mti}, CMX~\cite{zhang2023cmx}, and OGI-specific methods: Gasformer~\cite{sarker2024gasformer}, GasTwinFormer~\cite{tahamid2025gastwinformer}, and CarboFormer~\cite{islam2025carbonext}. Single-stream baselines receive 2-channel concatenated inputs (CO$_2$+CH$_4$); multimodal and OGI baselines use dual-stream configuration matching FUME.

All models are implemented in PyTorch and trained on NVIDIA A6000 GPU using AdamW optimizer (lr=0.001), batch size 16, 20 epochs with cosine annealing. Data augmentation includes random flipping, rotation ($\pm$15$^\circ$), and color jittering. Latency is measured with batch size 1, FP32 precision, CUDA synchronized timing (100 warmup + 1000 iterations), \texttt{cudnn.benchmark=True}. mIoU is computed per-modality across 3 classes and averaged over CO$_2$ and CH$_4$ heads.

\subsection{Evaluation Metrics}
\label{subsec:metrics}

\noindent\textbf{Classification:} Accuracy and per-class F1 scores (Macro F1 averaged across classes).

\noindent\textbf{Segmentation:} Mean IoU and Dice coefficient, where $\text{IoU} = |P \cap G|/|P \cup G|$ and $P$, $G$ denote predicted and ground-truth masks.

\noindent\textbf{Boundary Quality:} Hausdorff Distance 95\textsuperscript{th} percentile (HD95) \cite{kaur2022evolution} and Average Surface Distance (ASD) \cite{kaur2022evolution}; lower values indicate better boundary delineation.

\noindent\textbf{Efficiency:} Parameters (M), MACs (G), FPS, and GPU latency (ms).

\section{Results}
\label{sec:results}

\begin{figure*}[t]
\centering
\includegraphics[width=\textwidth]{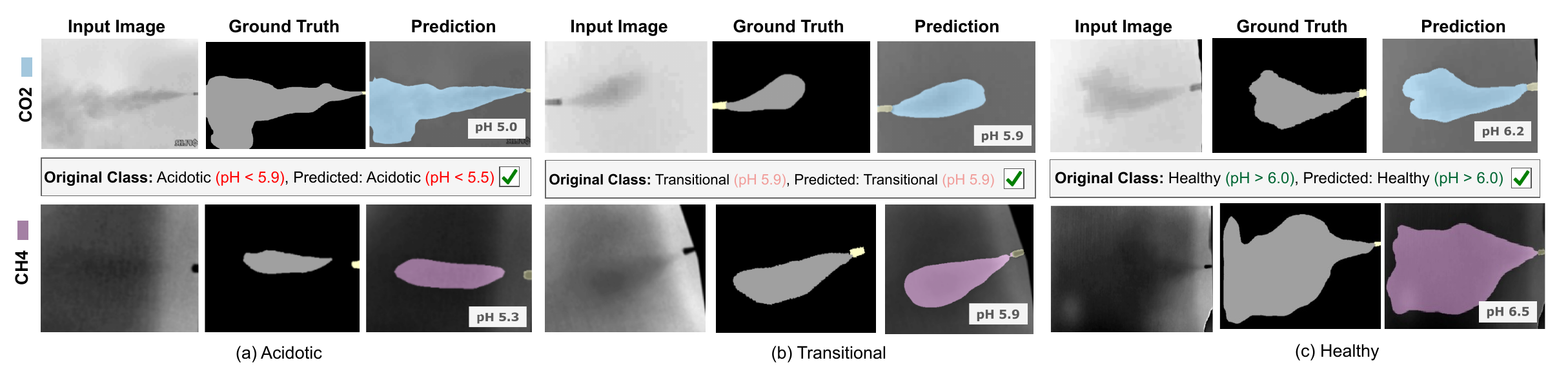}
\caption{Qualitative segmentation results across Acidotic, Transitional, and Healthy classes for both CO$_2$ and CH$_4$ modalities. FUME achieves precise boundary delineation of gas plumes with predictions closely matching ground truth annotations, demonstrating robustness to varying plume morphologies and emission intensities.}
\label{fig:qualitative_results}
\end{figure*}

\subsection{Comparison with State-of-the-Art Methods}
\label{subsec:comparison}

Table~\ref{tab:main_results} presents comprehensive quantitative comparisons including OGI-specific methods. FUME achieves the best segmentation quality with 80.99\% mIoU, outperforming both general segmentation baselines and OGI-specific methods: Gasformer~\cite{sarker2024gasformer} (62.09\%), CarboFormer~\cite{islam2025carbonext} (54.95\%), and GasTwinFormer~\cite{tahamid2025gastwinformer} (47.80\%). FUME also achieves the best boundary metrics (HD95: 46.58, ASD: 14.00) and high classification accuracy (98.82\%), substantially higher than OGI methods (86--90\%), while using fewer parameters (1.28M vs 3.3--3.8M) and lowest MACs (1.97G). At 326.80 FPS, FUME is 1.8--3$\times$ faster than OGI baselines, making it suitable for real-time edge deployment.

Fig.~\ref{fig:comparison_combined} provides visual comparison. The bar chart (A) illustrates the mIoU gap between methods, showing FUME achieves 80.99\% while others fall short by 5.08--54.30 pp. The radar chart (B) compares per-class F1 scores across all methods, demonstrating FUME's balanced performance across Healthy, Transitional, and Acidotic classes, while several baselines exhibit severe degradation on challenging classes. Detailed per-class metrics in Table~\ref{tab:perclass} show FUME achieves 97.49\% macro F1 and 99.03\% balanced accuracy, with the highest Acidotic class F1 (99.15\%), critical for acidosis detection.

\subsection{Efficiency Analysis}
\label{subsec:efficiency}

Fig.~\ref{fig:efficiency} visualizes the efficiency-quality trade-off. FUME occupies the optimal upper-left region (highest mIoU, minimal MACs). OGI-specific methods (Gasformer~\cite{sarker2024gasformer}, CarboFormer~\cite{islam2025carbonext}, GasTwinFormer~\cite{tahamid2025gastwinformer}) cluster in the mid-range with 3--5G MACs and 47--62\% mIoU, demonstrating that single-gas architectures struggle with dual-modality fusion. BiSeNetV2~\cite{yu2021bisenet} achieves good mIoU but requires 10$\times$ more computation. FUME uniquely achieves state-of-the-art segmentation at 326.80 FPS with lowest computational cost.

\subsection{Ablation Studies}
\label{subsec:ablation}

\begin{figure*}[t]
\centering
\includegraphics[width=0.85\textwidth]{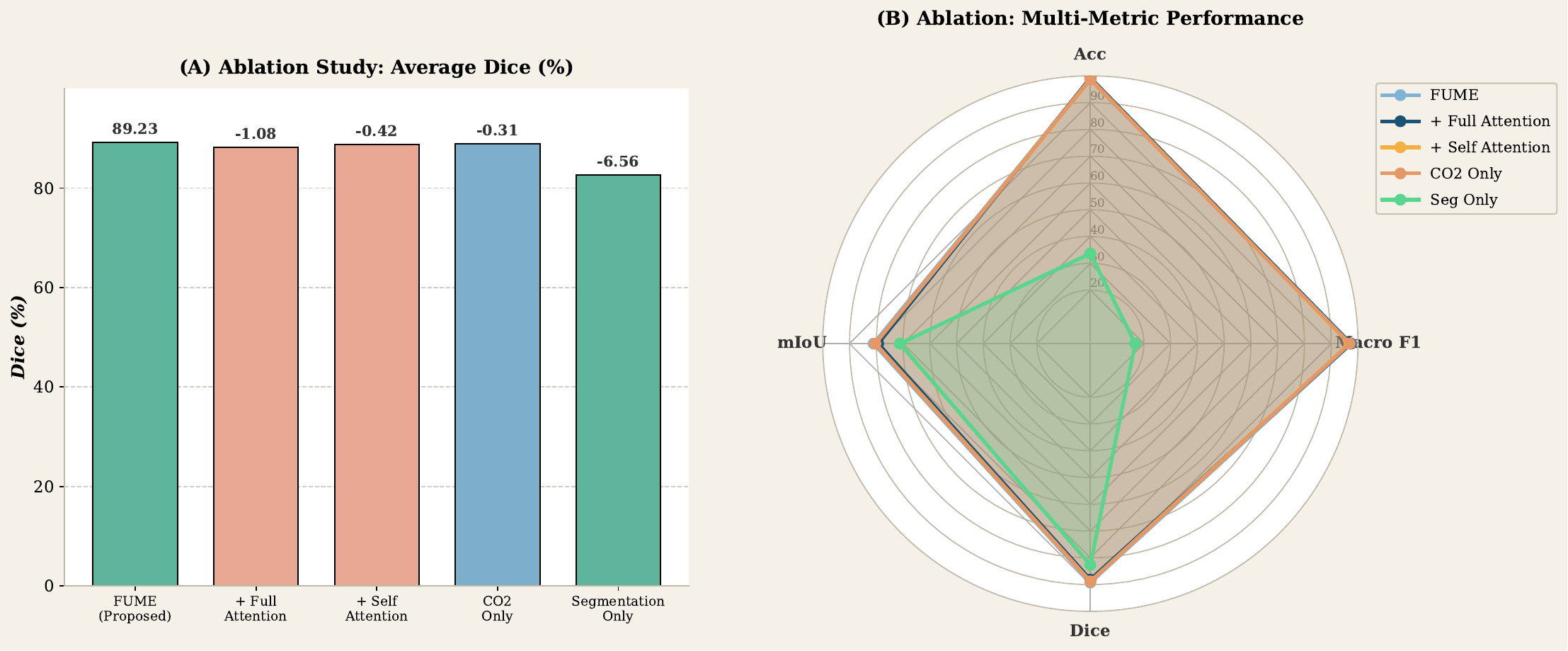}
\caption{Ablation study results. (A) Dice coefficient comparison across architectural variants, demonstrating that the proposed FUME configuration achieves optimal segmentation performance. (B) Multi-metric radar chart visualizing the critical importance of the CO$_2$ modality and dual-task learning framework for achieving balanced performance across accuracy, segmentation quality, and classification metrics.}
\label{fig:ablation_combined}
\end{figure*}

To validate our architectural design choices, we conduct systematic ablation studies. Table~\ref{tab:ablation} and Fig.~\ref{fig:ablation_combined} present the results of removing or modifying key components.

\noindent\textbf{Effect of Attention Mechanisms.} A counter-intuitive finding is that adding attention mechanisms—whether full cross-modal attention or self-attention only—degrades segmentation quality ($-1.68$ pp and $-0.63$ pp mIoU respectively) while significantly increasing inference latency (+48\% and +16\%). This challenges the common assumption that attention always benefits multi-modal fusion. We hypothesize that for tasks where modalities lack pixel-wise spatial correspondence, direct feature concatenation may provide sufficient alignment; however, this finding may be sensitive to attention design choices and hyperparameter tuning, warranting further investigation across different attention variants.

\noindent\textbf{Modality Contributions.} Each gas modality contributes distinct information: removing CO$_2$ (CH$_4$ Only) causes catastrophic classification failure (48.29\% accuracy), confirming CO$_2$ as the primary discriminative signal for acidosis detection. Conversely, removing CH$_4$ (CO$_2$ Only) degrades segmentation quality ($-0.47$ pp mIoU, $-0.31$ pp Dice), indicating that CH$_4$'s spatial diffusion patterns enhance boundary delineation. This asymmetric yet complementary relationship---CO$_2$ for discriminative power, CH$_4$ for spatial refinement---validates our dual-modality design.

\noindent\textbf{Task Interaction.} Both classification and segmentation branches are necessary for optimal performance. Using only the classification branch achieves high accuracy (97.86\%) but has no segmentation capability (no segmentation head), while the segmentation-only variant fails at classification (33.65\% accuracy) despite maintaining reasonable segmentation (71.21\% mIoU). This validates our dual-task learning framework where both tasks provide complementary supervisory signals.

\begin{table}[t]
\centering
\caption{Ablation study results. $\Delta$ denotes change relative to FUME.}
\label{tab:ablation}
\resizebox{\columnwidth}{!}{
\begin{tabular}{l|cccc|c}
\toprule
\textbf{Variant} & \textbf{Acc (\%)} & \textbf{mIoU (\%)} & \textbf{Dice (\%)} & \textbf{Latency} & \textbf{$\Delta$ mIoU} \\
\midrule
+ Full Cross-Modal Attn & 99.25 & 79.31 & 88.15 & 4.55 & $-1.68$ \\
+ Self-Attention Only & 98.82 & 80.36 & 88.81 & 3.55 & $-0.63$ \\
w/o CH$_4$ (CO$_2$ Only) & 98.61 & 80.52 & 88.92 & 1.78 & $-0.47$ \\
w/o CO$_2$ (CH$_4$ Only) & 48.29 & -- & -- & 1.81 & -- \\
Classification Only & 97.86 & -- & -- & 3.23 & -- \\
Segmentation Only & 33.65 & 71.21 & 82.67 & 2.95 & $-9.78$ \\
\midrule
\textbf{FUME (Proposed)} & \textbf{98.82} & \textbf{80.99} & \textbf{89.23} & \textbf{3.06} & \textbf{--} \\
\bottomrule
\end{tabular}
}
\end{table}

Fig.~\ref{fig:qualitative_results} shows qualitative results across all three classes. FUME achieves excellent boundary delineation across varying morphologies---irregular acidotic patterns, intermediate transitional states, and larger healthy regions---maintaining sharp plume boundaries even with diffuse emissions and low contrast.

\subsection{Discussion}
\label{subsec:discussion}

\noindent\textbf{Design Trade-offs.} Our results reveal a trade-off between classification accuracy and segmentation quality in attention-based fusion. Attention mechanisms boost classification (+0.43\% accuracy) but degrade segmentation ($-1.68$ pp mIoU). For applications requiring precise boundary delineation, FUME without attention provides superior performance.

\noindent\textbf{Modality Importance.} CO$_2$ and CH$_4$ encode physiologically distinct metabolic signals: CO$_2$ reflects fermentation rate while CH$_4$ indicates methanogenic activity, which is suppressed under acidotic conditions. Our ablation confirms this complementarity---CO$_2$-only achieves strong classification but CH$_4$ fusion improves segmentation quality (+0.47 pp mIoU, +0.31 pp Dice), particularly for boundary delineation where CH$_4$'s diffusion patterns provide additional spatial cues. Crucially, CH$_4$-only fails entirely (48.29\% accuracy), validating that both modalities contribute non-redundant information: CO$_2$ provides discriminative power while CH$_4$ refines spatial precision. This dual-gas framework also establishes a foundation for future multi-modal livestock monitoring systems.

\noindent\textbf{Limitations.} This study validates FUME under controlled in vitro conditions; translating to in vivo settings will require addressing animal motion, environmental variability, and intermittent plume visibility. Future work will explore temporal modeling, domain adaptation, and validation with live animals under farm conditions.

\section{Conclusion}
\label{sec:conclusion}

We presented FUME, a lightweight dual-stream architecture for rumen acidosis detection from dual-gas optical imaging that jointly performs gas plume segmentation and health classification through weight-shared encoders and channel attention fusion. Our experiments demonstrate state-of-the-art segmentation quality and classification accuracy with exceptional computational efficiency, outperforming both general segmentation baselines and OGI-specific methods. Key findings reveal that CO$_2$ serves as the primary biomarker for acidosis detection, dual-task learning provides essential complementary supervision, and channel attention fusion outperforms expensive cross-modal attention while maintaining real-time inference capability. Our work establishes proof-of-concept for gas emission-based livestock health monitoring, with future directions including in vivo validation and temporal modeling. The dataset and code will be released upon acceptance to facilitate future research.

\section{Acknowledgment}
\label{sec:acknowledgment}

This work is supported by the United States Department of Agriculture, National Institute of Food and Agriculture (USDA-NIFA), through the Capacity Building Grants for Non-Land-Grant Colleges of Agriculture (Grant No. 2023- 70001-40997).

{
    \small
    \bibliographystyle{ieeenat_fullname}
    \bibliography{main}
}

\end{document}